\documentclass[11pt,a4paper]{article}
\usepackage{times,latexsym}
\usepackage{url}
\usepackage[T1]{fontenc}
\usepackage{paralist}
\usepackage{graphicx}
\usepackage[utf8]{inputenc}
\usepackage{inconsolata}

\usepackage{zshi}
\usepackage{caption}
\usepackage{subcaption}
\usepackage{algorithm}
\usepackage[noend]{algorithmic}
\usepackage{microtype}

\newcommand\blfootnote[1]{%
  \begingroup
  \renewcommand\thefootnote{}\footnote{#1}%
  \addtocounter{footnote}{-1}%
  \endgroup
}

\usepackage[acceptedWithA]{tacl2021v1}
\usepackage{xspace,mfirstuc,tabulary}

\title{Red Teaming Language Model Detectors with Language Models}

\author{
Zhouxing Shi$^*$, Yihan Wang$^*$, Fan Yin$^*$, Xiangning Chen, Kai-Wei Chang, Cho-Jui Hsieh\\
University of California, Los Angeles\\
\texttt{\{zshi, yihanwang, fanyin20, xiangning, kwchang, chohsieh\}@cs.ucla.edu}\\
$^*$Alphabetical order
}
\date{}
\begin{document}
\maketitle

\begin{abstract}
The prevalence and strong capability of large language models (LLMs) present significant safety and ethical risks if exploited by malicious users. To prevent the potentially deceptive usage of LLMs, recent works have proposed algorithms to detect LLM-generated text and protect LLMs. In this paper, we investigate the robustness and reliability of these LLM detectors under adversarial attacks. We study two types of attack strategies: 1) replacing certain words in an LLM's output with their synonyms given the context; 2) automatically searching for an instructional prompt to alter the writing style of the generation. In both strategies, we leverage an auxiliary LLM to generate the word replacements or the instructional prompt. Different from previous works, we consider a challenging setting where the auxiliary LLM can also be protected by a detector. Experiments reveal that our attacks effectively compromise the performance of all detectors in the study with plausible generations, underscoring the urgent need to improve the robustness of LLM-generated text detection systems. 
\blfootnote{
Preprint. 
Accepted for publication at Transactions of the Association for Computational Linguistics (TACL) by MIT Press.  
Code will be released at: \url{https://github.com/shizhouxing/LLM-Detector-Robustness}.}
\end{abstract}

\section{Introduction}

Large language models (LLMs), such as ChatGPT~\citep{openai_chatgpt}, PaLM~\citep{chowdhery2022palm} and LLaMA~\cite{touvron2023llama}, have demonstrated human-like capabilities to generate high-quality text, follow instructions, and respond to user queries. Although LLMs can improve the work efficiency of humans, they also pose several ethical and safety concerns, such as it becomes hard to differentiate LLM-generated text from human-written text. For example, LLMs may be inappropriately used for academic plagiarism or creating misinformation at large scale~\citep{zellers2019defending}. Therefore, it is important to develop reliable approaches to protecting LLMs and detecting the presence of AI-generated texts, to mitigate the abuse of LLMs.

Toward this end, previous work has developed methods for automatically detecting text generated by LLMs. Existing methods mainly fall into three categories: 1) Classifier-based detectors by training a classifier, often a neural network, from data with AI-generated/human-written labels~\citep{solaiman2019release, AITextClassifier}; 2) Watermarking~\citep{kirchenbauer2023watermark} by injecting patterns into the generation of LLMs such that the pattern can be statistically detected but imperceptible to humans; 3) Likelihood-based detectors, e.g., DetectGPT~\citep{mitchell2023detectgpt}, by leveraging the log-likelihood of generated texts.  
However, as recent research demonstrates that text classifiers are vulnerable to adversarial attacks~\citep{iyyer2018adversarial,ribeiro2018semantically,alzantot2018generating}, these LLM text detectors may not be reliable when faced with adversarial manipulations of AI-generated texts. 

In this paper, we stress-test the reliability of LLM text detectors. We assume that there is an LLM $G$ that generates an output $\rmY=G(\rmX)$ given input $\rmX$. $G$ is \emph{protected} when there exists a detector $f$ that can detect text normally generated by $G$ with high accuracy. An \emph{attack} aims to manipulate the generation process such that a new output $\rmY'$ is still plausible given input $\rmX$ while the detector fails to identify $\rmY'$ as LLM-generated. The attack may leverage another attacker LLM $G'$.

In this context, we propose two novel attack methods.  In the first method, we prompt $G'$ to generate candidate substitutions of words in $\rmY$, and we then choose certain substitutions either in a query-free way or through a query-based evolutionary search~\citep{alzantot2018generating} to attack the detector. Our second method focuses on classifier-based detectors for instruction-tuned LLMs such as ChatGPT~\citep{openai_chatgpt}. We automatically search for an additional instructional prompt with a small subset of training data for a given classifier-based detector. At inference time, the additional instructional prompt instructs the LLM to generate new texts that are hard to detect.

Several concurrent studies~\citep{sadasivan2023can, krishna2023paraphrasing} proposed to attack detectors by paraphrasing AI-generated texts, with a different language model $G'$ for paraphrasing. However, they assume that $G'$ here is \textit{not protected} by a detector. Paraphrasing with $G'$ has been shown as effective in attacking detectors designed for the original LLM $G$, but it can become much less effective when $G'$ is also protected by a detector since the paraphrased model can still be detected by the detectors of $G'$. In contrast, we demonstrate that even when the attacker LLM $G'$ is also \textit{protected} by a detector, we can still leverage $G'$ for attacking LLM detectors. Therefore, even if all the strong LLMs are protected in the future, the currently existing detectors can still be vulnerable to our attacks.

We experiment with our attacks on all three aforementioned categories of LLM detectors. Our results reveal that all the tested detectors are vulnerable to our proposed attacks. The detection performance of these detectors degrades significantly under our attacks, while the texts produced by our attacks still mostly maintain reasonable quality as verified by human evaluation. Our findings suggest the current detectors are not sufficiently reliable yet and it requires further efforts to develop more robust LLM detectors.

\section{Related Work}
\label{sec:related_works}

\paragraph{Detectors for AI-generated text.}

Recent detectors for AI-generated text mostly fall into three categories.
First, classifier-based detectors are trained with labeled data to distinguish human-written text and AI-generated text. For example, the AI Text Classifier developed by OpenAI~\citep{AITextClassifier} is a fine-tuned language model. 
Second, watermarking methods introduce distinct patterns into AI-generated text, allowing for its identification. Among them, \citet{kirchenbauer2023watermark} randomly partition the vocabulary into a greenlist and a redlist during the generation, where the division is based on the hash of the previously generated tokens. The language model only uses words in the greenlists, and thereby the generated text has a different pattern compared to human-written text which does not consider such greenlists and redlists.
Third, DetectGPT~\citep{mitchell2023detectgpt} uses the likelihood of the generated text for the detection, as they find that text generated by language models tends to reside in the negative curvature region of the log probability function. Consequently, they define a curvature-based criterion for the detection.

\paragraph{Methods for red-teaming detectors.}

As the detectors emerge, several concurrent works showed that the detectors may be evaded to some extent, typically by paraphrasing the text~\citep{sadasivan2023can,krishna2023paraphrasing}. However, they need additional paraphrasing models which are typically unprotected models that are much weaker than the original LLM. Besides paraphrasing, \citet{kirchenbauer2023watermark} also discussed attacks against watermarking detectors with word substitutions generated by a masked language model such as T5 \citep{raffel2020exploring} which is a relatively weaker language model and tends to generate results with lower quality, and thus it may generate attacks with lower quality. On the other hand, \citet{chakraborty2023possibilities} analyzed the possibilities of the detection given sufficiently many samples.  

\paragraph{Adversarial examples in NLP.}
Word substitution is a commonly used strategy in generating textual adversarial examples~\citep{alzantot2018generating,ren2019generating,jin2020bert}.  Language models such as the BERT~\citep{devlin2018bert} have also been used for generating word substitutions~\citep{shi2020robustness_pi,li2020bert,garg2020bae}. In this work, we demonstrate the effectiveness of using the latest LLMs for generating high-quality word substitutions, and our query-based word substitutions are also inspired by the genetic algorithm in~\citet{alzantot2018generating,yin-etal-2020-robustness}. 
For our instructional prompt, it is relevant to recent works that prompt LLMs to red team LLMs themselves~\citep{perez2022red} rather than detectors in this work.
In addition, we fix a single instructional prompt at test time, which is partly similar to universal triggers in adversarial attacks~\citep{wallace2019universal,behjati2019universal}, but unlike them constructing an unnatural sequence of tokens as the trigger, our prompt is natural and it is added to the input for the generative model rather than the detector directly.

\paragraph{Safety of large language models.}
Detecting AI-generated texts is also related to the broader topic of LLM safety. Research for the safety of LLMs aims to reduce privacy leakage and intellectual property concerns~\citep{wallace2020imitation, carlini2021extracting, jagielski2023students, zhao2023protecting}, detect potential misuse~\citep{hendrycks2018deep, perez2022red}, defend against malicious users or trojan~\citep{wallace2019universal, wallace2021concealed}, or detecting hallucinations~\citep{zhou2021detecting,liu2022token}. See~\citet{hendrycks2021unsolved} for a roadmap of machine learning safety challenges. We test the reliability of some LLM text detection systems, which helps better understand the current progress in LLM text detection.

\section{Settings and Overview}
\label{sec:setting}

\begin{table*}[t]
\centering
\adjustbox{max width=\textwidth}{
\begin{tabular}{cc|ccc|ccc}
\toprule
\multicolumn{2}{c|}{\multirow{2}{*}{Attack}} & \multirow{2}{*}{Perturbation type} & \multicolumn{2}{c|}{Test-time Queries} & \multicolumn{3}{c}{Applicability}\\
&& & $G'$ & $f$ & Classifier & Watermarking & Likelihood\\
\midrule
Query-free &\multirow{2}{*}{Word Substitutions}& Output & \checkmark & - & \checkmark & \checkmark & \checkmark\\
Query-based&& Output & \checkmark & \checkmark & \checkmark & - & \checkmark\\
\multicolumn{2}{c|}{Instructional Prompts} & Input & - & - & \checkmark & - & - \\
\bottomrule
\end{tabular}}
\caption{
Properties of various attack methods  and their applicability to various detectors. ``Test-time queries'' indicates whether each method requires querying $G'$ or $f$ for multiple times at test time. 
}
\label{tab:methods}
\end{table*}

We consider a large language model $G$ that conditions on an input context or prompt $\rmX$ and generates an output text $\rmY=G(\rmX)$. We use upper-case characters to denote a sequence of tokens. For example, $\rmX=[\rvx_1, \rvx_2, ..., \rvx_m]$, where $m$ is the sequence length. The model $G$ is protected by a detector $f(\rmY) \in [0,1]$ that predicts whether $\rmY$ is generated by an LLM, where a higher detection score $f(\rmY)$ means that $\rmY$ is more likely to be LLM-generated. We use $\tau$ to denote a detection threshold such that $\rmY$ is considered LLM-generated if $f(\rmY)\geq\tau$. 

In this work, we consider three categories of detectors: (1) classifier-based detectors, (2) watermarking detectors, and (3) likelihood-based detectors. For classifier-based detectors, a text classifier $f(\rmY)$ is trained on a labeled dataset with $G$-generated and human-written texts. For watermarking detectors, $G$ is modified from a base generator $G_0$ with a watermarking mechanism $W$, denoted as $G=W(G_0)$, and a watermark detector $f(\rmY)$ is constructed to predict whether $\rmY$ is generated by the watermarked LLM $G$. Specifically, we consider the watermarking mechanism in \citet{kirchenbauer2023watermark}.
For likelihood-based detectors, they estimate $f(\rmY)$ by comparing the log probabilities of $\rmY$ and several random perturbations of $\rmY$. Specifically, we consider DetectGPT~\citep{mitchell2023detectgpt}. We consider a model $G$ as \textit{protected} if there is a detector $f(\rmY)$ in place to protect the model from inappropriate usage.

To stress test the reliability and robustness of those detectors in this setting, we develop red-teaming techniques to generate texts that can downgrade a detector using an LLM that is also protected by this detector. 
We consider attacks by output perturbation and input perturbation respectively:
\begin{compactitem}
\item \textbf{Output perturbation} perturbs the original output $\rmY$ and generates a perturbed output $\rmY'$.
\item \textbf{Input perturbation} perturbs the input $\rmX$ into $\rmX'$ as the new input, leading to a new output $\rmY'=G(\rmX')$.
\end{compactitem}
In both cases, we aim to minimize $f(\rmY')$ so that the new output $\rmY'$ is wrongly considered as human-written by the detector $f$. Meanwhile, we require that $\rmY'$ has a quality similar to $\rmY$ and remains a plausible output to the original input $\rmX$. 
For our attack algorithms, we also assume that the detector $f$ is black-box -- only the output scores are visible but not its internal parameters.

We propose to attack the detectors in two different ways. 
In \Cref{sec:word_replacement}, we construct an output perturbation by replacing some words in $\rmY$, where we prompt a protected LLM $G'$ to obtain candidate substitution words, and we then build query-based and query-free attacks respectively to decide substitution words.
In \Cref{sec:prompt_attack}, if $G$ is able to follow instructions, we search for an instructional prompt from the generation by $G$ and append the prompt to $\rmX$ as an input perturbation, where the instructional prompt instructs $G$ to generate texts in a style making it hard for the detector to detect.
\Cref{tab:methods} summarizes our methods and their applicability to different detectors.
At test time, instructional prompts are fixed and thus totally query-free. For word substitutions, they require querying $G'$ multiple times to generate word substitutions on each test example; the query-free version does not repeatedly query $f$ while the query-based version also requires querying $f$ multiple times. In practice, we may choose between these methods depending on the query budget and their applicability to the detectors.

\section{Attack with Word Substitutions}
\label{sec:word_replacement}

To attack the detectors with output perturbations, we aim to find a perturbed output $\rmY'$ that is out of the original detectable distribution. 
This is achieved by substituting certain words in $\rmY$. To obtain suitable substitution words for the tokens in $\rmY$ that preserve the fluency and semantic meaning, we utilize a protected LLM denoted as $G'$. For each token in $\rmY$ denoted as $\rvy_k$, we use $s(\rvy_k, \rmY, G', n)$ to denote the process of generating at most $n$ word substitution candidates for $\rvy_k$ given the context in $\rmY$ by prompting $G'$, and $s(\rvy_k, \rmY, G', n)$ outputs a set of at most $n$ words. Note that not every word can be substituted, and $s(\rvy_k, \rmY, G', n)$ can be an empty set if it is not suitable to replace $\rvy_k$. We will discuss how we generate the word substitution candidates using $G'$ in \Cref{sec:substitution_candidates}.

\paragraph{General attack objective.}
The objective of attacking $f$ with word substitutions can be formulated as a minimization problem given a substitution budget $\epsilon$:
\begin{align}
& \rmY' = \argmin_{\rmY'} f(\rmY'), \label{eq:minimization}\\
\text{s.t.} \quad & \rvy'_k \in \{\rvy_k\} \cup s(\rvy_k, \rmY, G', n) \nonumber,\\
& \sum_{k=1}^m \mathbbm{1}(\rvy_k \neq \rvy'_k) \leq \epsilon m. \nonumber
\end{align}
Here we aim to find an optimally perturbed output $\rmY'$ that minimizes the predicted score $f(\rmY')$ among all possible $\rmY'$. Each word in the perturbed output $\rvy_k'$ is either the unperturbed word $\rvy_k$ or selected from the word substitution candidates $s(\rvy_k, \rmY, G', n)$, and the total number of perturbed words is at most $\eps m$. 
To solve the minimization problem in \eqref{eq:minimization}, we consider both query-free and query-based substitutions respectively. We may choose between the two methods depending on whether the attacker can query $f$ for multiple times. 

\subsection{Generating Word Substitution Candidates}
\label{sec:substitution_candidates}

\begin{table}[t]
\center
\adjustbox{max width=.48\textwidth}{
\begin{tabular}{@{}m{\linewidth}@{}} 
\toprule
ChatGPT\\
\midrule
\textbf{Given this sentence: "}\emph{The scientists said the findings could help improve treatments and lead to new ways of preventing breast cancer spreading. Experts have described the findings as ``exciting''. }\textbf{", for each word in "}\emph{findings, exciting}\textbf{", give 10 substitution words that do not change the meaning of the sentence. Return each word and its substitutions in one line, in the format of "word:substitutions"}\\
\vspace{5pt}
{\color{blue}findings: results, discoveries, outcomes, conclusions, observations, data, evidence, findings, research, studies}\\
{\color{blue}exciting: thrilling, exhilarating, stimulating, electrifying, gripping, captivating, invigorating, inspiring, breathtaking, exciting}\\
\midrule
LLaMA\\
\midrule
\textbf{"}\emph{The scientists said the findings could help improve treatments and lead to new ways of preventing breast cancer spreading. Experts have described the findings as ``exciting''.}\textbf{"}\\
\textbf{Synonyms of the word} \emph{"exciting"} \textbf{in the above sentence are:}\\
\textbf{a)} {\color{blue}"interesting"}\\
{\color{blue} b) "surprising"}\\
{\color{blue}c) "unusual"}\\
\bottomrule
\end{tabular}}

\caption{Prompts for generating word substitution candidates using ChatGPT and LLaMA  as well as the corresponding outputs.
Text in bold denotes the prompt template.
Text in italic denotes a text to be perturbed or words to be replaced for a given example.
The generated word substitutions are in blue and listed after the bold text.
}
\label{tab:word_substituion_format}
\end{table}

\Cref{tab:word_substituion_format} shows the prompts we use and the outputs produced by $G'$, when $G'$ is ChatGPT and LLaMA respectively. ChatGPT is able to follow instructions, and thus our prompt is an instruction asking the model to generate substitution words, and multiple words can be substituted simultaneously. For LLaMA which has less instruction-following ability, we expect it to generate a text completion following our prompt, where the prompt is designed such that a plausible text completion consists of suggested substitution words, and we replace one word at a time.

The benefit of applying an LLM here is that it enables us to obtain substitution words that not only have similar meanings with the original word but are also compatible with the context, as previous works also used language models such as BERT for generating adversarial examples~\citep{shi2020robustness_pi}. Thus it is more convenient than earlier methods using synonym lists for generating substitution words which need to be further checked with a separate language model~\citep{alzantot2018generating} for compatibility with the context.

\subsection{Query-based Word Substitutions}
\label{sec:query_based}
For query-based substitutions, we use the evolutionary search algorithm~\citep{alzantot2018generating, yin-etal-2020-robustness} originally designed for generating adversarial examples in NLP.
The algorithm starts from a \textit{population} of perturbed texts which includes input texts with a certain amount of tokens randomly replaced. Then, it iterates over several \textit{generations} of populations to select elites in each population, i.e, the most effective substitution that leads to the lowest detection score. New generations are constructed by crossing over the elite substitutions in the previous generation.

\subsection{Query-free Word Substitutions}
\label{sec:query_free}
For the query-free attack, we simply apply word substitution on random tokens in $\rmY$ to attack DetectGPT and classifier-based detectors. 
For watermarking detectors, we further design an effective query-free attack utilizing the properties of the detection method. 

Specifically, for the watermarking in \citet{kirchenbauer2023watermark}, the watermarked LLM generates a token by modifying the predicted logits at position $i+1$: $g(\rvy_{i+1}|[\rvy_1, ..., \rvy_i]) = g_0(\rvy_{i+1}|[\rvy_1, ..., \rvy_i]) + \delta$ if the candidate token $\rvy_{i+1}$ is in the greenlist, where we use $g_0$ to denote the output logits of the original model $G_0$ and $g$ for the watermarked model $G$. 
$\delta$ is an offset value for shifting the logits of greenlist tokens, and $\gamma$ is the proportion of greenlist tokens in the vocabulary. 
Therefore, a text generated by the watermarked model tends to have more greenlist tokens compared to a text generated by the original model.
 $f(\rmY)$ calculates the detection score based on the number of greenlist tokens in $\rmY$ as:
\begin{align}
    f(\rmY) = (|s_G| - \gamma T)/\sqrt{T\gamma(1-\gamma)},
    \label{eq:watermark}
\end{align}
where $|s_G|$ is the number of greenlist tokens in $\rmY$ and $T$ is the total number of tokens in $\rmY$.

Therefore, given a fixed substitution budget $\epsilon$, we aim to identify and substitute more greenlist tokens to reduce the total count of greenlist tokens. We achieve this with a two-stage algorithm. At the first stage, we sort all tokens in $\rmY$ by the prediction entropy estimated with a language model $M$, which can be either the same generative model $G$ or a weaker model as we only use the entropy as a heuristic score. 
The prediction entropy is estimated by feeding $M$ with the prefix or masked text without the word to be estimated.
As the watermarking offset $\delta$ is applied on the decoding process, a token with higher entropy is easier to be affected by watermarking. At the second stage, we pick $\epsilon m$ tokens with highest entropy and use a watermarked LLM $G'$ to generate word substitutions as introduced in Section \ref{sec:substitution_candidates}.

\begin{table*}[ht]
\center
\adjustbox{max width=.75\textwidth}{
\begin{tabular}{cccc}
\toprule
Generative Model & DetectGPT& Watermarking& Classifier-based Detector \\
\midrule
GPT-2-XL & ChatGPT& LLaMA-65B& ChatGPT  \\
LLaMA-65B & LLaMA-65B& LLaMA-65B& -  \\
ChatGPT & ChatGPT& - &  ChatGPT  \\
\bottomrule
\end{tabular}}
\caption{The protected LLM $G'$ used in generating perturbations for each generative model $G$ and the detectors. ``-'' indicates a combination of the generative model and the detector is not applicable.
}
\label{tab:models}
\end{table*}

\section{Attack by Instructional Prompts}
\label{sec:prompt_attack}

In this section, we build attacks by perturbing the input prompt to encourage LLMs to generate texts that are difficult to detect. In particular, we focus on LLM-based generative models that can follow instructions and classifier-based detectors. We consider ChatGPT~\citep{openai_chatgpt} as the generative model $G$ and OpenAI AI Text Classifier~\citep{AITextClassifier} as the detector $f$.
The OpenAI AI Text Classifier is a fine-tuned neural network, while neural networks have been shown to be vulnerable to distribution shifts in NLP literature~\citep{miller2020effect,awadalla2022exploring}. 
Therefore, we aim to shift the generated text to a different distribution where the detector is more likely to fail.
We do not require the shifted generation to be semantically equivalent to the original text, but the generation should still be a plausible output to the given input.

We achieve this by searching for an additional prompt $\rmX_p$ appended to the original input $\rmX$, which forms a new input $\rmX'=[\rmX, \rmX_p]$ to $G$. In particular, $\rmX_p$ consists of $\rmX_{\text{ins}}$ and $\rmX_{\text{ref}}$, where $\rmX_{\text{ins}}$ is an instruction asking the model to follow the writing style of reference $\rmX_{\text{ref}}$. 

\paragraph{Searching for $\rmX_p$.}

We search for $\rmX_p$ on a small subset of training examples with $n$ examples $\rmX_1, \rmX_2, \cdots, \rmX_n$. We assume that we can query the detector $f$ for multiple times during search time. After an effective $\rmX_p$ is found, it can be applied universally on all inputs from this dataset at test time. 
The objective of the search is:
\begin{equation}
\argmin_{\rmX_p} \frac{1}{n} \sum_{i=1}^n \mathbbm{1}(f(G([\rmX_i, \rmX_p]))\geq \tau),
\label{eq:prompt_obj}
\end{equation}
which aims to minimize the average detection rate for the new outputs generated with $\rmX_p$ appended to the input.

We use $G$ to generate various $\rmX_{\text{ins}}$ and $\rmX_{\text{ref}}$ in each iteration and try to search for an optimal $\rmX_p=[\rmX_{\text{ins}},\rmX_{\text{ref}}]$ following the objective in \eqref{eq:prompt_obj}.
Initially, we set $\rmX_{\text{ins}}$ to a manually written instruction, ``Meanwhile please imitate the writing style and wording of the following passage:''. An initial value for $\rmX_{\text{ref}}$ is not necessary.
We also create and initialize a priority queue $\gO$ with $n$ initial outputs generated from the $n$ training examples without $\rmX_p$. $\gO$ sorts its elements according to the detection scores from $f$ and prioritize those with lower scores.
In each iteration of the search, we have two steps:
\begin{compactitem}
\item Updating $\rmX_{\text{ref}}$:  We pop the top-$K$ candidates from $\gO$. For each candidate, we combine it with the current  $\rmX_{\text{ins}}$ respectively as the potential candidates for $\rmX_p$ in the current iteration.
\item Updating $\rmX_{\text{ins}}$: We instruct model $G$ to generate $K$ variations of the current $\rmX_{\text{ins}}$, inspired by \citet{zhou2022large} for automatic prompt engineering. And we combine them with the current $\rmX_{\text{ins}}$ respectively as the potential candidates for $\rmX_p$.
\end{compactitem}
For both of these two steps, we take the best candidate $\rmX_p$ according to \eqref{eq:prompt_obj}.
When generating $G([\rmX_i,\rmX_p])$ in \eqref{eq:prompt_obj}, we push all the generated outputs to $\gO$ as the candidates for $\rmX_{\text{ref}}$ in the later rounds.
We take $T$ iterations and return the final $\rmX_p = [\rmX_{\text{ins}}, \rmX_{\text{ref}}]$ to be used at test time, and $\rmY'=G([\rmX, \rmX_p])$ is the new output given input $\rmX$. 

For some $\rmX_p$, we find that the $G$ may directly copy text from $\rmX_{\text{ref}}$ to generate $\rmY'$ when $\rmX_p$ is appended into the input prompt. To prevent this behavior, we compute a matching score between $\rmX_{\text{ref}}$ and  $\rmY$ and discard a candidate $\rmX_p$ during the search if more than 20\% of words from $\rmX_{\text{ref}}$ (except stop words) appear in $\rmY'$. In this way, we find that the copying behavior is effective prevented.

\section{Experiments}
\begin{table*}[ht]
\center
\scalebox{0.8}{\begin{tabular}{cccccccccccccc}
\toprule
Generative Model & Dataset & Unattacked & Dipper Paraphrasing & Query-free Substitution &Query-based Substitution\\
\midrule
\multirow{2}{*}{GPT-2-XL} & XSum & 84.4 & 35.2 & 25.9 & \textbf{3.9} \cr
& ELI5 & 70.6& 36.7 & 21.2 & \bf 3.8 \cr
\midrule
\multirow{2}{*}{ChatGPT} & XSum & 56.0& 34.6& 25.6 &\bf 4.5\cr
& ELI5 & 55.0 & 39.5 & 12.2 &\bf  6.5 \cr
\midrule
\multirow{2}{*}{LLaMA-65B} & XSum & 59.3& 49.0& 25.5 & \textbf{9.9}\cr
& ELI5 & 60.5 & 53.1 & 31.4 & \textbf{18.6}\cr
\bottomrule
\end{tabular}}
\caption{AUROC scores (\%) of DetectGPT under various attack settings.}
\label{tab:detectgpt_exp}
\end{table*}
\begin{table*}[ht]
\center
\adjustbox{max width=.8\textwidth}{
\begin{tabular}{ccccccccccccc}
\toprule
\multirow{2}{*}{Generative Model} & \multirow{2}{*}{$\delta$} & \multirow{2}{*}{Dataset} &\multicolumn{2}{c}{Unattacked} & \multicolumn{2}{c}{Dipper Paraphrasing} & \multicolumn{2}{c}{Query-free Substitution}\\
& & & AUROC & DR & AUROC & DR &AUROC & DR\\
\midrule
\multirow{4}{*}{GPT-2-XL} & \multirow{2}{*}{1.0}& XSum&98.0&81.0&\textbf{84.7}&38.0&85.9&\textbf{34.0}\\
& & ELI5 & 97.14 & 72.0 & \textbf{86.3} & 34.0 & 86.9 & \textbf{33.0}\\
& \multirow{2}{*}{1.5}& XSum&99.4&96.0&94.1&72.0&\textbf{91.6}&\textbf{35.0}\\
& & ELI5 & 98.7 & 91.0 & 95.0 & 67.0 & \textbf{91.5} & \textbf{47.0}\\
\hline
\multirow{4}{*}{LLaMA-65B} &\multirow{2}{*}{1.0}& XSum & 88.9 & 22.0 &79.3&12.0&\textbf{67.8} & \textbf{10.0}\\
&& ELI5 & 94.7 & 70.0 & 81.6 & 45.0 & \textbf{79.6}& \textbf{34.0}\\
&\multirow{2}{*}{1.5}& XSum & 96.4 & 63.0 &85.5& 27.0&\textbf{84.5} & \textbf{16.0}\\
&& ELI5 & 99.6 & 95.0 & 92.3 & 72.0 & \textbf{91.8} & \textbf{61.0}\\
\bottomrule
\end{tabular}}
\caption{Attack against watermarking detector. We report both AUROC scores (\%) and the detection rates (DR) (\%).
For DR, we set the decision threshold such that the false positive rate for the human reference text on the same test examples  is 1\%.}
\label{tab:watermark_exp}
\end{table*}

\subsection{Experimental Settings}

\paragraph{Generative Models and Detectors.} 
We experiment with a wide range of generative LLMs and corresponding detectors. For the generative model $G$, we consider GPT-2-XL~\citep{radford2019language}, LLaMA-65B~\citep{touvron2023llama}, and ChatGPT (\texttt{gpt-3.5-turbo}) ~\citep{openai_chatgpt}.
For detectors, we consider DetectGPT~\citep{mitchell2023detectgpt}, watermarking~\citep{kirchenbauer2023watermark}, and classifier-based detectors~\citep{AITextClassifier, solaiman2019release}. 
For DetectGPT, we use GPT-Neo~\citep{gpt-neo} as the scoring model to estimate the log-likelihood.
DetectGPT also requires masking spans of the texts and filling in the spans with an external T5-3B model~\citep{raffel2020exploring}. We fix the mask rate to be 15\%. 
Watermarking is applied to open-source LLaMA-65B and GPT-2-XL but not ChatGPT, as it requires logits scores in generation. We use $\gamma=0.5$ in all the watermarking experiments, following the default setting in \citet{kirchenbauer2023watermark}. 
Moreover, classifier-based detectors include a fine-tuned RoBERTa-Large detector~\citep{solaiman2019release} for GPT-2 texts and the OpenAI AI Text Classifier~\citep{AITextClassifier} for ChatGPT texts. We summarize all generative models and detectors considered in the experiments in Table \ref{tab:models}.

For experiments involving watermarking, we use a watermarked LLaMA-65B as $G'$, as we cannot implement watermarking on ChatGPT; we also use LLaMA-65B as $G'$ in the setting with LLaMA-65B itself as $G$ and DetectGPT as the detector; and in other settings, we use ChatGPT as $G'$ which is protected by either DetectGPT or the classifier-based detector. The choice of protected LLM $G'$ is also summarized in \Cref{tab:models}.

\paragraph{Baselines.}
To demonstrate the advantage of our methods in revealing detectors' weakness, we compare with several baselines. Dipper paraphrase~\citep{krishna2023paraphrasing} is a recent method that trains a paraphrasing model to rewrite AI texts and bypass detectors. It prepends diversity codes to control the level of paraphrases introduced to the texts. We use \texttt{Dipper-paraphraser-XXL} with 20 lexical diversity and 60 order diversity to paraphrase the AI texts, which keeps the same level of $20\%$ uni-gram difference as word substitution. We use nucleus sampling~\citep{holtzman2019curious} with $p=0.9$ for the paraphraser. And in ChatGPT experiments, we also use ChatGPT itself for paraphrasing. 

\paragraph{Datasets.}
We mainly use two types of datasets including text completion and long-form question answering. 
We use XSum~\citep{Narayan2018DontGM} for text completion, where we take the first sentence as the input prompt for the completion, and we use ELI5~\citep{eli5_lfqa} for long-form question answering. 
In addition, for the RoBERTa-Large detector, we also use a specific GPT-2 output dataset~\citep{solaiman2019release} as they are fine-tuned solely on GPT-2 texts. 
Since the OpenAI AI Text Classifier requires the text to contain at least 1,000 characters, we filter XSum and ELI5 datasets and only retain examples with human reference text containing at least 1,000 characters. For each dataset, we shuffle the test set and use the first 100 examples.

\paragraph{Metrics.}
We use several metrics to evaluate the detectors under attacks. Area Under the Receiver Operating Characteristic Curve (AUROC) scores summarize the performance of detectors under various thresholds. A detection rate (DR) is the true positive rate under a fixed threshold (positive examples mean LLM-generated texts), where we either tune the threshold to meet a particular false positive rate or follow the original thresholds of the detectors. For the GPT-2 output dataset, we also use Attack Success Rate (ASR) which computes the rate that the attack successfully flips the prediction by the detector, out of all the positive examples on which the detector originally predicts correctly. 
\begin{table}[ht]
\centering
\adjustbox{max width=.46\textwidth}{
\begin{tabular}{@{}m{0.52\textwidth}@{}}
\toprule
Original output \\
\hline
The scientists said the findings could help improve treatments and lead to new ways of preventing breast cancer spreading. Experts have described the findings as ``exciting''. Bone is the most common site for breast cancer to spread to. Once breast cancer reaches the bone, it can be treated but often is not curable. In experiments in mice, the Sheffield researchers found breast cancer cells were sending signals to the cells inside\\
\hline
Query-free Substitution\\
\hline
The \textbf{researchers} said the findings could help improve \textbf{drugs} and lead to new ways of \textbf{stopping} breast cancer \textbf{metastasizing}. Experts have \textbf{said} the discoveries as ``\textbf{interesting}''. Bone is the most common \textbf{place} for \textbf{bone} cancer to \textbf{metastasize} to. Once \textbf{bone} cancer \textbf{spreads} \textbf{to} the bone, it can be treated but usually is not \textbf{can be treated}. In \textbf{research} in \textbf{the Sheffield researchers}, the Sheffield \textbf{Scientists} found \textbf{cancer} cancer cells were sending signals to the cells \textbf{in}.\\
\bottomrule
\end{tabular}}
\caption{An example from the XSum dataset. We show the original output from watermarked LLaMA-65B, as well as the output after query-free word substitution attack.
}
\label{watermark_example}
\end{table}

\subsection{Attack with Word Substitutions}

We apply word substitution-based attack on all  the three categories of detection methods. In each setting, we assume that both $G$ and $G'$ are protected by the same detector $f$. 
We limit the number of substituted words to be at most 20\% of the total number words except stop words and proper nouns that should not be substituted. This leads to around 7 substituted tokens per example. For the evolutionary search, it requires 70 queries per example with a population size of 10. 

\begin{figure*}[!htb]
    \centering
    \begin{minipage}[t]{0.45\textwidth}
        \centering
        \includegraphics[width=\linewidth]{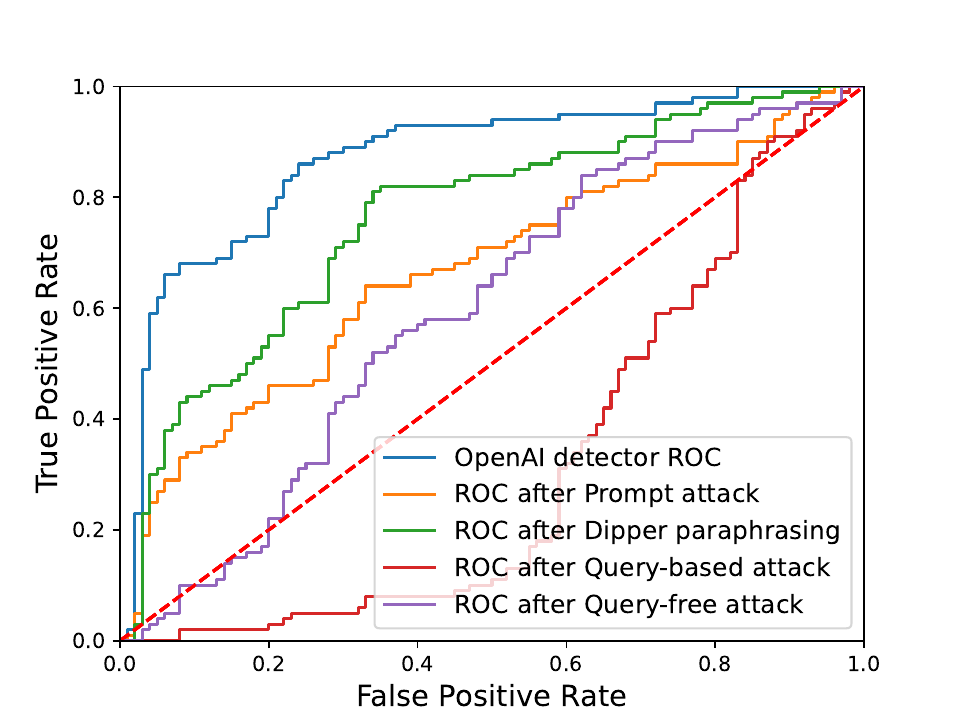}
        \caption{ROC plot of OpenAI AI Text Classifier under different attack methods. We show the ROC plot on the ELI5 dataset in \Cref{tab:openai_detector}.}
        \label{fig:openai_detector_auc}
    \end{minipage}%
    \hfill
    \begin{minipage}[t]{0.45\textwidth}
        \centering
        \includegraphics[width=\linewidth]{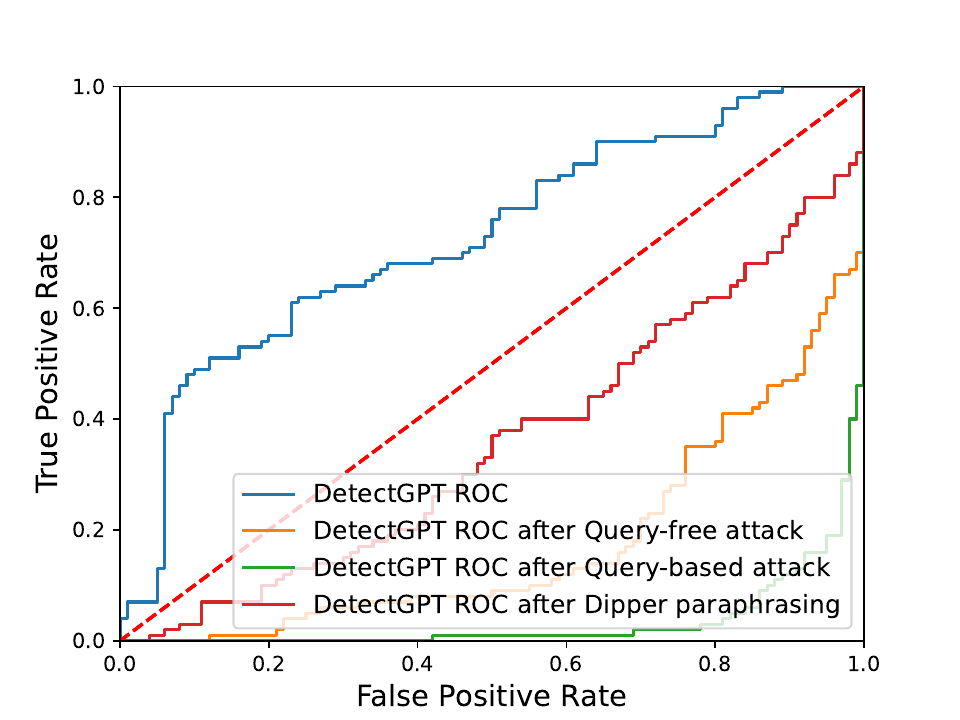}
        \caption{ROC plot of DetectGPT detectors under different attack methods. We show the ROC plot on the ELI5 dataset in \Cref{tab:detectgpt_exp} for the GPT2-XL model.}
        \label{fig:detectGPT_auc}
    \end{minipage}\\
    \begin{minipage}[t]{0.45\textwidth}
        \centering
        \includegraphics[width=\linewidth]{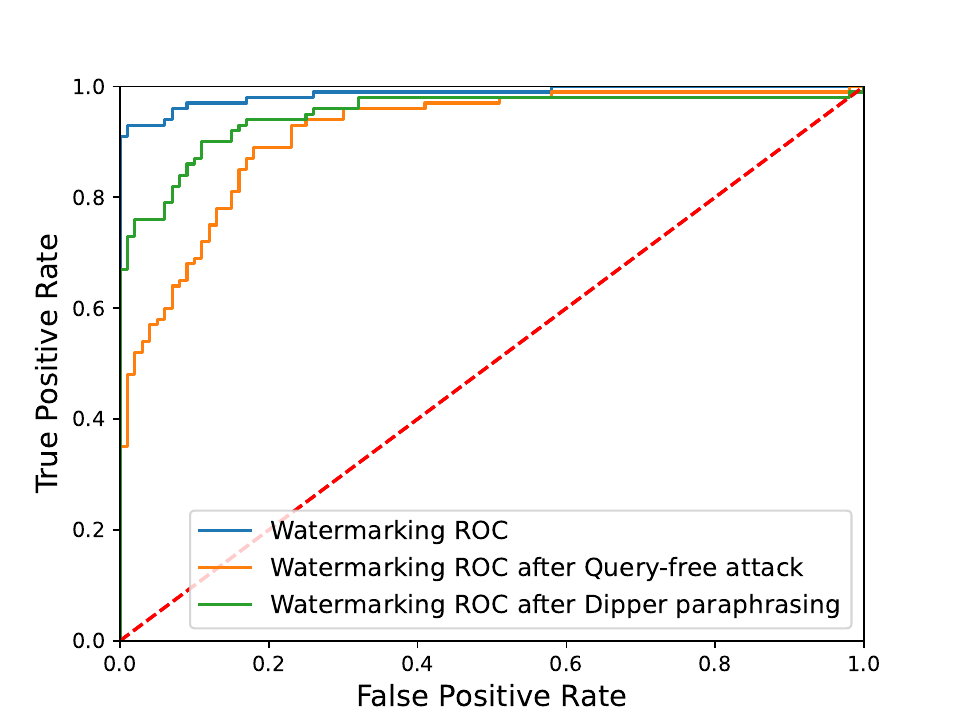}
        \caption{ROC plot of watermarking detectors under different attack methods. We show the ROC plot on the ELI5 dataset in \Cref{tab:watermark_exp} for the GPT2-XL model with $\delta=1.5$.}
        \label{fig:watermarking_auc_gpt2}
    \end{minipage}
    \hfill
    \begin{minipage}[t]{0.45\textwidth}
        \centering
        \includegraphics[width=\linewidth]{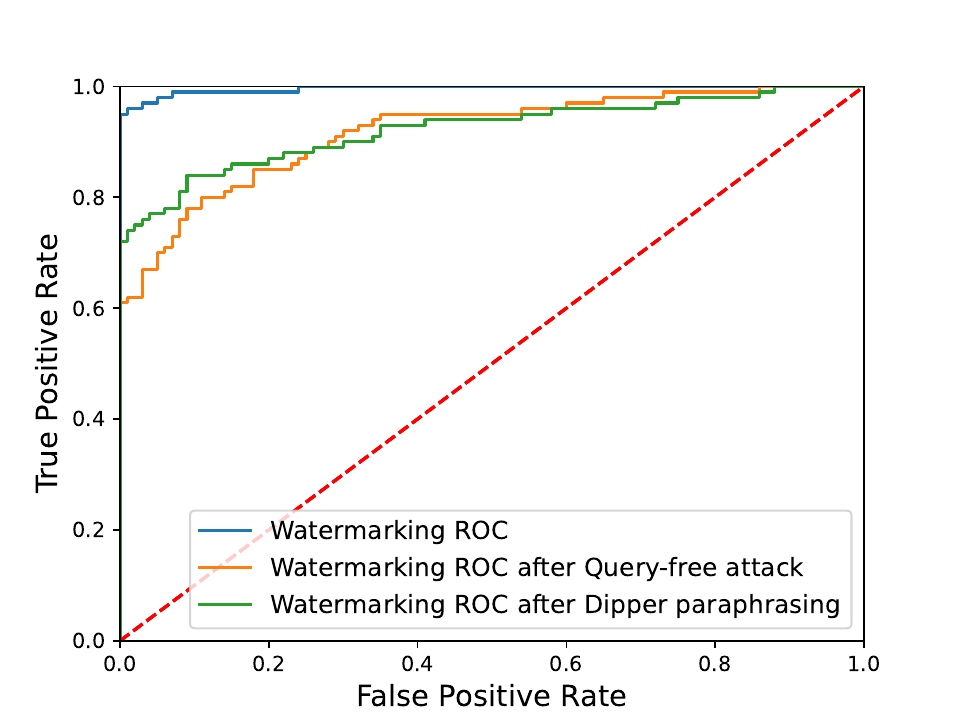}
        \caption{ROC plot of watermarking detectors under different attack methods. We show the ROC plot on the ELI5 dataset in \Cref{tab:watermark_exp} for the LLaMA-65B model with $\delta=1.5$.}
        \label{fig:watermarking_auc_llama}
    \end{minipage}\\
    \begin{minipage}[t]{0.45\textwidth}
        \centering
        \includegraphics[width=\linewidth]{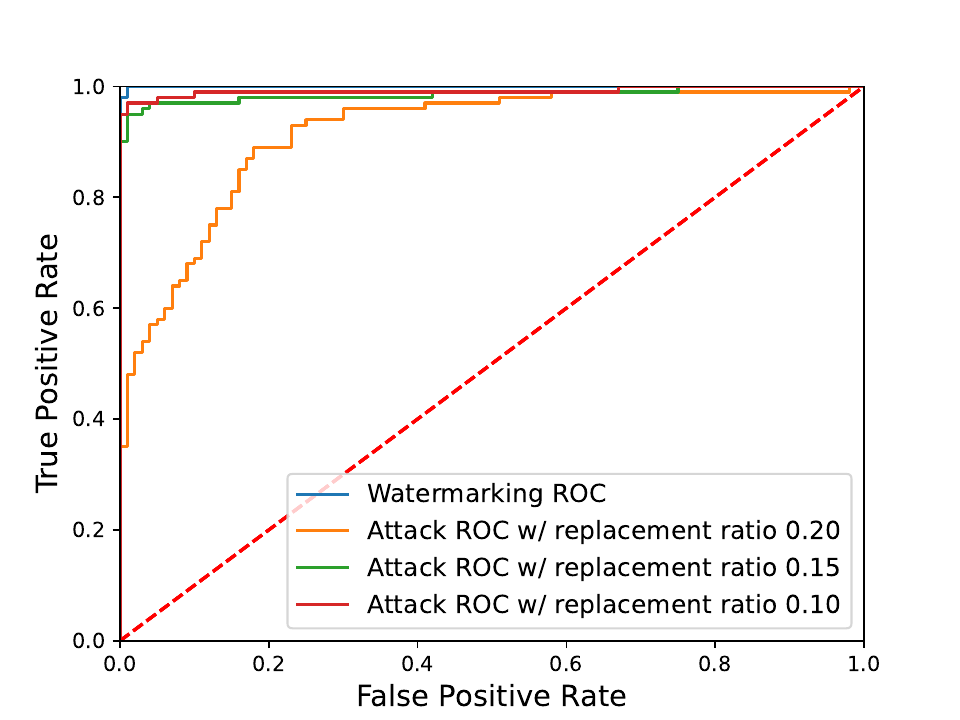}
        \caption{ROC plot of watermarking detectors under query-free attacks with different replacement ratios. We show the ROC plot on the ELI5 dataset for the GPT2-XL model with $\delta=1.5$.}
        \label{fig:gpt2_ratio}
    \end{minipage}
    \hfill
    \begin{minipage}[t]{0.45\textwidth}
        \centering
        \includegraphics[width=\linewidth]{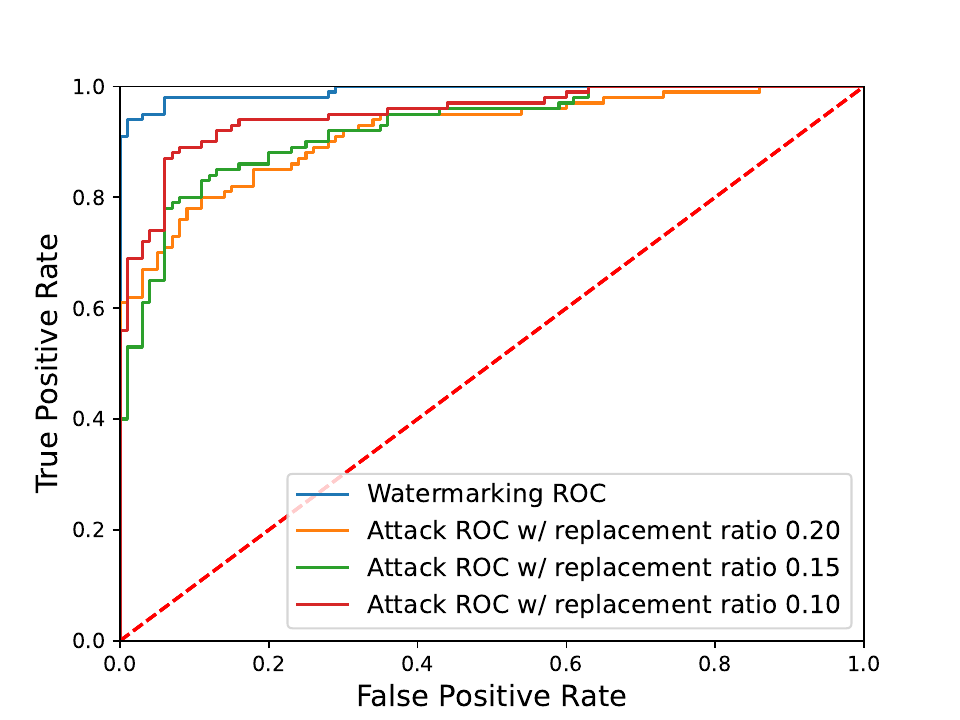}
        \caption{ROC plot of watermarking detectors under query-free attacks with different replacement ratios. We show the ROC plot on the ELI5 dataset for the LLaMA-65B model with $\delta=1.5$.}
        \label{fig:llama_ratio}
    \end{minipage}
\end{figure*}

\paragraph{Attack against DetectGPT} 
The results are shown in Table \ref{tab:detectgpt_exp}. We show an example of ROC plot in Figure \ref{fig:detectGPT_auc}.
We find that DetectGPT's AUROC drops below random guess when we simply apply query-free substitutions which randomly select substitutions from the candidate pool. For example, on XSum, AUROC on GPT-2-XL drops from 84.4\% to 25.9\%. For ChatGPT and LLAMA-65B, AUROC drops from 56.0\% to 25.6\% and from 59.3\% to 25.5\%, respectively. 
The AUROC scores further drop to only 3.9\%, 4.5\%, and 9.9\% respectively with the query-based evolutionary search. Our word substitution methods consistently surpass the Dipper paraphrasing, which demonstrates that our methods are revealing more vulnerability of the detectors. 
We do not use DR here as we find that the DR values are already very low  (usually below 10\%) even when no attack is applied, and we follow \citet{mitchell2023detectgpt} which originally also only used AUROC.
\paragraph{Attack against Watermarking}

We use a T5-Large model to estimate the prediction entropy for each token. In our main experiments, we select 20\% of tokens in the initial output $\rmY$ with highest prediction entropy to be replaced in the attack. We also compare the attack performance with different replacement ratios in Figure \ref{fig:gpt2_ratio} and Figure \ref{fig:llama_ratio}. In all experiments in Table \ref{tab:watermark_exp}, we keep the first 100 tokens in each text. We filter the suggested word substitutions to keep fewer than 4 tokens in the substitution candidates to avoid invalid substitution. We report the AUROC score and detection rate for each setting in \Cref{tab:watermark_exp}. We show two examples of ROC plot in Figure \ref{fig:watermarking_auc_gpt2} and \ref{fig:watermarking_auc_llama}. For the detection rates, we set the threshold value to keep the false positive rate for human texts equal to 1\% following \citet{krishna2023paraphrasing}. The results show that the detection rates are significantly degraded after the query-free word substitution attack under two different watermarking settings with $\delta=1.0, 1.5$. Although the detection rate on unattacked texts can be further increased by increasing $\delta$, in practice, the watermark strength should be kept under an appropriate level to avoid hurting the quality of text generation \citep{kirchenbauer2023watermark}.
Compared to Dipper paraphrasing, we achieve lower detection rates without using a separate unprotected paraphraser model. 
We also show qualitative examples for attacks against watermarking in \Cref{watermark_example}.

\paragraph{Attack against Classifier-based Detectors}

\begin{table}[ht]
\centering
\adjustbox{max width=.32\textwidth}{
\begin{tabular}{ccc}
\toprule
\multicolumn{2}{c}{Attack} & ASR\\
\midrule
Dipper Paraphrasing & &4\%\\
Query-free Substitutions & &2\%\\
Query-based Substitutions & &\bf 68\%\\
\bottomrule
\end{tabular}
}
\caption{Attack Success Rate (ASR) for OpenAI RoBERTa-Large detector for GPT-2 texts.}
\label{tab:gpt2}
\end{table}

Results for attacking GPT-2 text detector are shown in \Cref{tab:gpt2}. We find that the attack success rate (ASR) on detecting GPT-2 texts is close to 0 for both paraphrasing and query-free substitutions. We hypothesize that this is because the detector is specifically trained on detecting GPT-2 texts, and it is hard to remove the patterns leveraged by those detectors by randomly selecting word substitutions or paraphrasing. Our evolutionary search-based substitutions achieve much better ASR compared to the query-free methods. 

For the OpenAI AI Text Classifier shown in \Cref{tab:openai_detector}, query-free attacks are able to decrease the detection AUROC by 18.9 and 28.1 percentage points on XSum and ELI5, respectively, while query-based ones further decrease them by 45.4 and 55.6 percentage points to lower than random. Comparison with the attack using instructional prompts and more details are discussed in Section \ref{exp:promp_attack}.

\begin{table}[t]
\centering
\adjustbox{max width=.48\textwidth}{
\begin{tabular}{m{.50\textwidth}}
\toprule
Initial prompt for querying ChatGPT on XSum\\
\hline
{
Please complete this passage with at least 150 words:

\{$\rmX$\}
}\\
\midrule
Initial prompt for querying ChatGPT on ELI5\\
\hline
{
Please answer this question with at least 150 words:

\{$\rmX$\}
}\\
\midrule
Prompt for paraphrasing \\
\hline
{
Please paraphrase the following passage, with at least 200 words:

\{$\rmY$\}
}\\
\bottomrule
\end{tabular}}    
\caption{Prompts used for querying ChatGPT.
Initial prompts are used for instructing ChatGPT to perform text completion or question answering on XSum and ELI5 respectively. And the prompt for paraphrasing is used in \Cref{tab:openai_detector} for paraphrasing $\rmY$ into $\rmY'$ directly. We also instruct ChatGPT to generate at least 150 words as the OpenAI AI Text Classifier does not accept shorter texts.
}
\label{tab:chatgpt_prompts}
\end{table}

\begin{table}[ht]
\centering
\adjustbox{max width=.45\textwidth}{
\begin{tabular}{m{.50\textwidth}}
\toprule
$\rmX_p$ on XSum \\
\hline 
{
During the waiting period, please take into consideration utilizing the writing style and vocabulary used in the subsequent paragraph.

"Wales football star, Gareth Bale, is set to undergo surgery on his ankle after suffering an injury during Real Madrid’s 2-1 victory over Sporting Lisbon in the Champions League. (...) "
}\\
\midrule
$\rmX_p$ on ELI5 \\
\hline
{
At the same time, kindly mimic the writing technique and diction utilized in the subsequent excerpt.

"The reason why metal feels cooler compared to other things at the same temperature is due to its thermal conductivity. (...)"
}\\
\bottomrule
\end{tabular}}    
\caption{Our searched instructional prompts on XSum and ELI5 respectively. Part of the $\rmX_{\text{ref}}$ is omitted due to the space limit.}
\label{tab:instructional_prompts}
\end{table}

\begin{table}[ht]
\centering
\adjustbox{max width=.48\textwidth}{
\begin{tabular}{cccccc}
\toprule
\multirow{2}{*}{Method} & \multicolumn{2}{c}{XSum} & \multicolumn{2}{c}{ELI5}\\
& AUROC & DR & AUROC & DR \\
\midrule
Unattacked & 88.8 & 30.0 & 87.1 & 54.0 \\
ChatGPT Paraphrasing & 80.0 & 14.0 & 76.2 & 27.0 \\
\hline
Query-free Substitution & 69.9& 2.0 & 59.0 & 2.0\\
Query-based Substitution & \textbf{43.4} & \textbf{0.0} & \textbf{31.5} & \textbf{0.0}\\
Instructional Prompts & 54.9 & 5.0 & 66.7 & 21.0\\
\bottomrule
\end{tabular}}
\caption{
AUROC scores (\%) and detection rates (DR) (\%) of the OpenAI AI Text Classifier on the original outputs by ChatGPT and outputs with various attacks respectively. 
}
\label{tab:openai_detector}
\end{table}

\begin{table}[ht]
\centering
\adjustbox{max width=.48\textwidth}{
\begin{tabular}{cccc}
\toprule
Dataset & Method & Fluency & Plausibility \\
\midrule
\multirow{5}{*}{XSum} & Unattacked & 2.79$\pm$0.47 & 2.76$\pm$0.52 \\
& ChatGPT Paraphrasing & 2.60$\pm$0.52 & 2.75$\pm$0.43 \\
& Query-free Substitution & 2.58$\pm$0.57 & 2.68$\pm$0.50\\
& Query-based Substitution & 2.49$\pm$0.67 & 2.68$\pm$0.52 \\
& Instructional Prompts & 2.67$\pm$0.57 & 2.66$\pm$0.55\\
\hline
\multirow{5}{*}{ELI5} & Unattacked &  2.78$\pm$0.45 & 2.87$\pm$0.34 \\
& ChatGPT Paraphrasing  & 2.35$\pm$0.48 & 2.95$\pm$0.28 \\
& Query-free Substitution & 2.83$\pm$0.37 & 2.60$\pm$0.49 \\
& Query-based Substitution & 2.65$\pm$0.54 & 2.63$\pm$0.47 \\
& Instructional Prompts & 2.73$\pm$0.50 & 2.56$\pm$0.62\\
\bottomrule
\end{tabular}}
\caption{Average score and standard deviation of ratings from human evaluation on attacks against the OpenAI AI Text Classifier for ChatGPT.}
\label{tab:human_chatgpt}
\end{table}

\begin{table}[ht]
\centering
\adjustbox{max width=.48\textwidth}{
\begin{tabular}{cccc}
\toprule
Dataset & Method & Fluency & Plausibility \\
\midrule
\multirow{3}{*}{XSum} & Unattacked & 2.79$\pm$0.47 & 2.76$\pm$0.52 \\
& Dipper Paraphrasing & 2.67$\pm$0.51 & 2.60$\pm$0.52 \\
& Query-free Substitution & 2.58$\pm$0.57 & 2.68$\pm$0.50\\
& Query-based Substitution & 2.35$\pm$0.51 & 2.40$\pm$0.49 \\
\hline
\multirow{3}{*}{ELI5} & Unattacked &  2.78$\pm$0.45 & 2.87$\pm$0.34 \\
& Dipper Paraphrasing  & 2.63$\pm$0.48 & 2.68$\pm$0.47 \\
& Query-free Substitution & 2.83$\pm$0.37 & 2.60$\pm$0.49 \\
& Query-based Substitution & 2.47$\pm$0.50 & 2.42$\pm$0.53 \\
\bottomrule
\end{tabular}}
\caption{Average score and standard deviation of ratings from human evaluation on
attacks against DetectGPT for detecting ChatGPT generation.}
\label{tab:human_detectgpt}
\end{table}

\begin{table}[ht]
\centering
\adjustbox{max width=.48\textwidth}{
\begin{tabular}{cccc}
\toprule
Dataset & Method & Fluency & Plausibility \\
\midrule
\multirow{3}{*}{XSum} & Unattacked & 2.65$\pm$0.51 & 2.68$\pm$0.53 \\
& Dipper Paraphrasing & 2.68$\pm$0.47 & 2.58$\pm$0.59 \\
& Query-free Substitution & 2.45$\pm$0.69 & 2.68$\pm$0.47 \\
\hline
\multirow{3}{*}{ELI5} & Unattacked & 2.45$\pm$0.50 & 2.42$\pm$0.49\\
& Dipper Paraphrasing  &  2.63$\pm$0.51 & 2.66$\pm$0.53 \\
& Query-free Substitution & 2.27$\pm$0.62 & 2.58$\pm$0.49 \\
\bottomrule
\end{tabular}}
\caption{Average score and standard deviation of ratings from human evaluation on attacks against the watermarking detector for the watermarked LLaMA-65B with $\delta=1.0$, $\gamma=0.5$.}
\label{tab:human_watermark}
\end{table}
\begin{table*}[t]
\centering
\adjustbox{max width=\textwidth}{
\begin{tabular}{m{1.2\textwidth}}
\toprule
Original input \\
\hline
{
Please answer this question with at least 150 words:

Why we don't yet have a cable provider that will let me pick and pay for ONLY the channels I want?
}
\\
\hline
Original output\\
\hline
{
The reason why we do not yet have a cable provider that will allow choosing and paying for only the channels we want is that it may not be economically feasible for the providers. The cable providers typically negotiate content deals with media companies in packages or bundles. This packaging provides providers with a lower cost per channel, making it more profitable for them. (...)
}\\
\midrule
ChatGPT Paraphrasing \\
\hline
{
The lack of cable providers that offer the option to choose and pay for individual channels is due to the feasibility of the idea. Cable providers, such as media companies, negotiate bundles or packages of channels rather than individual offerings. This arrangement allows for a lower cost per channel for the providers, which translates to higher profits. 
Additionally, media companies benefit from bundling channels as it increases the exposure of lower-demand channels and ensures programming visibility. 
(...)
}\\
\midrule
Query-free Substitution\\
\hline
The \textbf{motive} why we do not yet have a cable provider that will allow choosing and paying for only the channels we want is that it may not be \textbf{monetarily} feasible for the providers. The cable providers typically negotiate content deals with media companies in \textbf{wrap} or bundles. This packaging provides providers with a lower \textbf{expense} per channel, making it more profitable for them. .
The media companies also benefit from \textbf{integrating} their channels, as they can sell lower-demand channels as a package deal, ensuring broader visibility for their programs. 
(...)\\
\midrule
Query-based Substitution \\
\hline
The reason why we do not yet have a cable provider that will allow \textbf{nominating} and paying for only the channels we want is that it may not be economically feasible for the providers. The \textbf{Cablegram} providers typically negotiate content deals with media companies in packages or bundles. This \textbf{parcel} provides providers with a lower cost per channel, making it more profitable for them. 
The media companies also benefit from packaging their channels, as they can sell lower-demand \textbf{avenue} as a package deal, ensuring broader visibility for their programs. 
(...)
\\
\midrule
Instructional Prompts \\ 
\hline
{
The reason why we don't yet have a cable provider that lets us pick and pay for only the channels we want is due to the complex and structured industry model that cable companies have established. The traditional cable model focuses on bundling channels together into packages, often forcing customers to pay for channels they don't want or need.
This model benefits cable companies by allowing them to earn more revenue from customers who are willing to pay for premium packages, even if they don't watch all the channels that come with it. 
(...)
}\\
\bottomrule
\end{tabular}}
\caption{
An example from the ELI5 dataset when using ChatGPT as the generative model. We show the original input and output, as well as the output under various attacks. Due to space limit, we omit part of the generation as indicated by ``(...)''.
}
\label{tab:example_eli5}
\end{table*}

\subsection{Attack with Instructional Prompts}
\label{exp:promp_attack}

We conduct experiments for our instructional prompts using ChatGPT as the generative model and the OpenAI AI Text Classifier as the classifier-based detector.
The detector is \texttt{model-detect-v2} accessible via OpenAI APIs as of early July, 2023. 
We choose this detector as it is developed by a relatively renowned company and has been shown to achieve stronger detection accuracy~\citep{krishna2023paraphrasing} than other classifier-based detectors such as GPTZero~\citep{tian2023gptzero}. This detector was also available at no cost when our experiments were conducted. 
Its output contains five classes, including ``likely'', ``possibly'', ``unclear if it is'', ``unlikely'' and ``very unlikely'', with thresholds 0.98, 0.90, 0.45, and 0.10 respectively. We follow these thresholds and use a threshold of 0.9 to compute detection rates.  

We search for the instructional prompt using $n=50$ training examples, $T=5$ iterations, and $K=5$ candidates in each iteration.
We show the prompts for querying ChatGPT in \Cref{tab:chatgpt_prompts} and the results in 
\Cref{tab:openai_detector}.
Our instructional prompts significantly reduce the the AUROC scores and detection rates compared to the unattacked setting and are more effective than paraphrasing with ChatGPT. 
While using instructional prompts may not lead to lower AUROC or DR compared to word substitutions, it does not require querying $G'$ or $f$ multiple times, making it a more efficient and equally effective option.
We show an example on ELI5 with various attacks in \Cref{tab:example_eli5} and the instructional prompts found by our algorithm in \Cref{tab:instructional_prompts}.

\section{Human Evaluation}

To validate that our approach mostly preserves the quality of the generated text, we conduct a human evaluation on Amazon Mechanical Turk (MTurk). On each dataset, we consider the first 20 test examples and ask 3 MTurk workers to rate the quality of the text generated by each method on each of the test examples. Specifically, we use two metrics, including fluency and plausibility, where fluency measures whether the text is grammatically correct and fluent, and plausibility measures whether the generated text is a plausible output given the input, on either the text completion (XSum) or long-form question answering (ELI5) task. We use a 1/2/3 rating scale for each of the metrics (3 is the best and vice versa), and we provide the workers with guidance on the ratings, according to whether there are many/several/almost no issues for the 1/2/3 ratings on fluency and plausibility respectively. The workers are paid USD \$0.05 for each example and we provide an additional bonus. The annotation time varies, but the estimated wage rate is \$10/hr, which is higher than the US minimum wage (\$7.25/hr).

\Cref{tab:human_chatgpt,tab:human_detectgpt,tab:human_watermark} show results on attacks against the three detectors respectively. These results show that our attack methods can maintain reasonable and satisfactory plausibility and fluency with a small degradation compared to the unattacked texts. Among our attack methods, we find that the query-free substitution usually has better fluency and also sometimes better plausibility compared to the query-based substitution, as the query-based one which aims to search for a stronger attack tends to degrade the text quality slightly more. 
Our method with instructional prompts has better fluency than the query-based substitution and sometimes better fluency than the query-free substitution, and its generation is directly from model $G$ without further substituting words; it also has comparable plausibility compared to the word substitution methods.

\section{Discussions}
\paragraph{Robustness of detectors.}
Comparing the results in \Cref{tab:detectgpt_exp,tab:watermark_exp,tab:openai_detector}, we see a clear trend that watermarking is relatively more robust to the attacks compared to the other two techniques. The detection mechanism in watermarking is mainly a statistical method and tends to be more robust compared to the likelihood-based and classifier-based detectors which heavily rely on neural networks. However, watermarking is also the only method here that modifies the generation process of the protected LM. It requires access to the intermediate outputs of the LM and the generation quality may degrade. While the three detectors are not strictly comparable, as watermarking has a different setting by modifying the generation, our results still show insights on the different degrees of robustness of the various detectors.
From \Cref{tab:detectgpt_exp} and \Cref{tab:openai_detector}, we can also see that query-based method generally produces stronger attacks, which benefits from the guidance of multiple queries to the detector when searching for more effective word substitutions.

\paragraph{Text quality under the attacks.}
We note that attack approaches often result in a minor decline in generation performance. Nonetheless, based on human assessment, the quality remains acceptable and is adequate to spam the target in a real-world scenario. 
Note that in practical scenarios such as online spamming, malicious actors do not have to use perfect text and they may still use text with slightly degraded quality, as their main purpose is not to generate perfect text but text that is hard to detect. 
Therefore, the insufficient robustness of existing detection strategies continues to be a significant concern.

\paragraph{Defending against the attacks.}
Inspired by our attack results, we discuss on potential directions for developing methods to defend against the attacks.
One possibility is to combine watermarking with a likelihood estimation to defend against word substitution attacks.
This is based on the observation that the word substitution attacks often need to substitute around 20\% tokens from greenlisted tokens to redlisted tokens.
After the word substitution,  the new redlisted tokens tend to have lower probabilities in the prediction by the original watermarked model, and the new text also tends to have a higher perplexity under the watermarked model. 
Thus, one may leverage a watermarked language model to check the perplexity or the likelihood of all the redlist tokens, to predict whether a word substitution attack is possibly present.

\section{Conclusion and Limitations}

In this work, we study the reliability of three distinct types of LLM text detectors by proposing two attack strategies: 1) word substitutions and 2) instructional prompts using protected LLMs. Experiments reveal the vulnerability of existing detectors, which urges the design of more reliable LLM text detectors.
We will release the source code and data with BSD-3-Clause license at GitHub upon acceptance.

Finally, the purpose of this work is to test and reveal the limitations of the currently existing LLM text detectors, and we red-team the detectors for future works to improve their robustness and reliability based on our proposed evaluation. Thus this work is potentially beneficial to for developing future systems protecting LLMs and preventing abusive usage. 
The proposed approaches should not be used to bypass real-world LLM text detectors.

\section*{Acknowledgements}

We thank UCLA-NLP, the action editor, and the reviewers for their invaluable feedback. The work is supported in part by CISCO, NSF 2008173, 2048280, 2325121, 2331966, ONR N00014-23-1-2300:P00001, and DARPA ANSR	FA8750-23-2-0004.

\bibliography{custom} 
\bibliographystyle{acl_natbib}
  
\end{document}